# Building a Parallel Corpus and Training Translation Models Between Luganda and English

Richard Kimera, Daniela N. Rim, Heeyoul Choi

***Abstract:*** Neural machine translation (NMT) has achieved great successes with large datasets, so NMT is more premised on high-resource languages. This continuously underpins the low resource languages such as Luganda due to the lack of high-quality parallel corpora, so even 'Google translate' does not serve Luganda at the time of this writing. In this paper, we build a parallel corpus with 41,070 pairwise sentences for Luganda and English which is based on three different open-sourced corpora. Then, we train NMT models with hyper-parameter search on the dataset. Experiments gave us a BLEU score of 21.28 from Luganda to English and 17.47 from English to Luganda. Some translation examples show high quality of the translation. We believe that our model is the first Luganda-English NMT model. The bilingual dataset we built will be available to the public.

## 1. Introduction

Uganda has not had an official language survey since 1971, and most scholars and reports have provided varying statistics about languages in Uganda. It is reported that there are 43 living indigenous languages [1] down from 63 varieties identified in 1972 [2].

Even with this diversity, the constitution provides for English as the official language and Swahili widely spoken in East Africa as the second language [3]. English is common among the elites, and is the language of instruction in schools and offices, serving as a unifying language in the country. However, Swahili remains largely expressed in the border districts and is not a de-facto language in the parliament or courts of law in Uganda [1]. Instead, Luganda is steadily positioning itself as an undeclared national language in Uganda with the last published report of 1998 showing a total of 4,130,000 speakers [4] out of a population of 22 million people. Recent reports indicate that over 20 million people can speak the language due to its influence in the business domain [5].

With the recent increase in the need for technology to deliver and access services online, communication is a catalyst. This calls for the use of language to pass on information between the involved parties. A case in point is using web portals to apply for and receive services like passports, visas, business, and tax registration. An assessment of 78 Uganda's E-government websites called for an improvement in making the websites multilingual [6]. These are usually delivered in an official language, English, that is not well understood by the Illiterate. Machine Translation (MT) has the power to utilize Natural Language Processing (NLP) models [7, 8, 9, 10] to translate these sites into the local languages.

However, since the recent NLP models need a large dataset as well as enough computation power to train the model on the datasets, it has been challenging to utilize NLP models in Uganda for technological advancements and service delivery due to the absence of a corpus, especially for the Luganda language. Furthermore, as of $1^{st}$ May 2022, platforms like Google translate, and IBM Watson have not yet incorporated Luganda into their translation systems. Moreso, the current NLP research that is done on the Luganda language has mainly focused on building small datasets [11, 12, 13].

In this work, we build a bilingual parallel corpus for Luganda and English by combining the various existing open datasets. We describe the data acquisition and preprocessing steps with the source of the datasets. We also apply the Transformer [10] for Neural Machine Translation (NMT) using the built dataset, which confirms the quality of the dataset. We train the model architectures with many different hyper-parameter configurations and the best one is obtained by the Bayesian approach that automates the hyper-parameter search.

In experiments, we achieved a BLEU score of 21.28 from Luganda to English and 17.47 from English to Luganda on the test data with 1,233 pairwise sentences. Also, we present some translation examples to show the quality of the translation. To the best of our knowledge, this is the first NMT models between Luganda and English, and we will open our bilingual dataset to the public.

## 2. Background

In this section, we present the key models used in MT and why we chose to use the Transformer.

### 2.1. Neural Machine Translation (NMT)

Inspired by the human brain, there are several types of Neural Networks (NNs) including feedforward, recurrent and convolutional networks [14]. They can further be classified as Deep Neural Networks

(DNNs) with an increase in the number of layers [7]. DNNs have made NMT the mainstream approach and technology to MT systems [15, 16, 17].

An NMT model is trained on a parallel text corpus. Using the Encoder-Decoder (E-D) architecture, a sentence from the source language is translated into a fixed-length vector by the encoder network which is then decoded to the target language [8].

Recurrent Neural Networks (RNNs) are one of the architectural choices for NMT, since they can process sequential data [7, 8, 9]. One of the setbacks of RNNs is the vanishing gradient, whereby longer sentences tend to deteriorate the performance of the E-D architecture even with Long Short-Term Memory (LSTM) [9, 18, 19]. Although not as popular as RNNs, Convolutional Neural Networks (CNN) were also used in NMT [20, 21, 22].

## 2.2. Transformer

Transformer was proposed to replace the previous network architectures like RNNs and CNNs in language processing [10]. Self-attention and feed-forward connections used together made the Transformer architecture a key model for advancing the field of neural machine translation. It increased the efficiency and reduced the speed of convergence. Transformer also supports parallel processing, hence leveraging the power of modern GPUs [10].

Generally, the Transformer architecture consists of encoder and decoder blocks. The encoder receives an input sequence $(x^{(1)}, x^{(2)}, \ldots, x^{(n_x)})$ and processes it, and the decoder generates an output sequence $(y^{(1)}, \ldots, y^{(n_y)})$, one element at a time [9]. The encoder and decoder consist of N stacked layers that include a multi-head attention and feed-forward layer. The decoder adds extra attention layer for the output of the encoder. It is finally connected to the linear and SoftMax layers to output the probabilities of the target language based on the input.

The Transformer has been widely utilized in high-resource languages like English, French, and German. The original paper utilized datasets of 4.5M (English to German) and 36M (English to French) sentence pairs [10]. Such a large dataset is not yet existent for the Luganda language; however, it is possible to utilize a smaller dataset as low as 30,000 as evidenced in English – isiXhosa translation [23]. In this paper, we use a dataset (Luganda-English pairwise corpus) focused on the use of Transformer for translation.

## 3. Luganda Datasets and NMT Training

Even though MT has been highly embraced, its utilization on the Luganda language has been a challenge due to the absence of the required quality of corpora. A many-to-many multilingual translation for African languages, excluding Luganda, was built with a mixture of biblical [24] and non-biblical corpora [25]. The training was based on a pre-trained variant of the transformer-based architecture mT5 model [26]. To solve the dataset challenge, as one of their goals, the Masakhane society [27] brings together NLP researchers in Africa to spur research in African languages. They focus on using a more participatory approach.

There has been an effort to build a model using a Luganda monolingual dataset for text to speech [28]. This research was limited to the usage of a small corpus. Even though the dataset was publicly published, it was not helpful in our research due to its monolingual nature. A community-based method was used to create a corpus of five based Ugandan languages (Ateso, Luganda, Lugbara, and Runyankole) [29]. Religious datasets have equally been utilized by web scrapping [30] text from websites (watchtower), magazines (awake), and the Bible [24, 31, 32].

An MT model to translate from Lumasaaba to English was trained on a Bible-based text corpus [33]. Lumaasaba is a closely related language to Luganda, however, the use of the Bible corpus is underscored as the language context does not usually represent the local contextual use [29].

As far as we know, there have been small bilingual datasets published [11, 12, 13], which calls for the expansion of their work. Additionally, NMT models had not yet been tried on any of the datasets.

3.1. Data Acquisition

We sourced and built a corpus from research centers; three different corpora were merged, and the quality of the translation was checked.

Corpus 1: This dataset was published in 2022 by Zenodo in collaboration with a team of researchers from Makerere AI and data science research lab, Luganda teachers, students, and freelancers. It consisted of a total of 1,042 English and Luganda parallel sentences [12]. The original text was downloaded as a csv file format, and text sentences have been extracted as seen in Table 1.

| English | Luganda |
| --- | --- |
| The teacher taught us how to multiply numbers yesterday. | Omusomesa yatusomesezza okukubisa emiwendo eggulo. |
| The doctor asked to see me physically because he couldn't diagnose me without a proper checkup. | Ddokita yasaba okundaba mu buliwo kubanga yali tasobola kumanya kinnuma nga takoze kukebera kutuufu. |

Table 1. Sample sentences from Corpus 1

Corpus 2: This dataset consists of 15,022 English-Luganda parallel sentences published in 2021. We show some sample text from the dataset as seen in Table 2. The corpus was built using the same procedure as Corpus 1 [11], which was originally a csv file.

| English | Luganda |
| --- | --- |
| Refugees had misunderstandings between themselves. | Abanoonyiboobubudamu b'abadde n'obutakkaanya wakati waabwe. |
| We were urged to welcome refugees into our communities. | Twakubirizibwa okwaniriza abanoonyiboobubudamu mu bitundu byaffe. |

Table 2. Sample sentences from Corpus 2

Corpus 3: This dataset had a total of 25,006 language phrases and was released in 2021 by sunbird AI in collaboration with the Makerere AI lab. The dataset consisted of a parallel text in English with corresponding translations in 5 local languages (Luganda, Lugbara, Runyakitara, Acholi, and Ateso). The English text was sourced from social media, transcripts from radio, online newspapers, articles, blogs, text contributions from Makerere University NLP community, and farmer responses from surveys [34]. The JSON format corpus was downloaded from the company's GitHub account [13]. Table 3 shows a set of the sampled text in the JSON format from the dataset.

Given the three datasets in csv or JSON file formats, we merged them into an xlsx format. The new created dataset was cleaned and formed a corpus of 41,070 pairwise sentences. We then proceeded with data preprocessing which consisted of splitting and tokenization. The corpus was split into training, validation, and testing portions of 94% (38,641), 3%, and 3%, respectively. This was done for each language and a total of 6 sets were generated. After tokenizing each set, we used Byte-Pair Encoding (BPE) to create a vocabulary of 10,000 words that were later indexed [35].

| |
| --- |
| "{"English":"Eggplants always grow best under warm conditions.", <br> "Luganda":"Bbiringanya lubeerera asinga kukulira mu mbeera ya bugumu", |

```
"Runyankole":"Entonga buriijo zikurira omu mbeera y'obwire erikutagata",
"Ateso":"Epoloi ebirinyanyi ojok apakio nu emwanar akwap.",
"Lugbara":"Birinyanya eyi zo kililiru ndeni angu driza ma alia.",
"Acholi":"Bilinyanya pol kare dongo maber ka lyeto tye"}"
```

Table 3. One sample from Corpus 3 in the JSON format with six languages: English, Luganda, Runyankole, Ateso, Lugbara, and Acholi

We further went ahead to confirm the quality of the corpus manually by checking that the translation was of a high standard. In this NMT task, we implemented a Transformer using the PyTorch library, the scripts were written in both Python and Perl. The experiments were run on an NVIDIA GeForce RTX 2080 Ti GPU on top of a Linux-powered OS.

### 3.2. Training

We trained an English-Luganda (En2Lu) model and a Luganda-English (Lu2En) model. During the training process, Adam [36] was used as the optimizer, and the early stopping trick was implemented to prevent the model from overfitting. We applied the trained models to check for performance using the test datasets. The experiments were conducted on a vocabulary of 10,000 words, with 6 layers of the encoder and 6 layers of the decoder [37].

To find the best hyper-parameters, we interfaced with weights and biases (wandb) [38], an MLOps online platform, we were able to keep track of our experiments as well as automate the hyper-parameter search. We sampled hyper-parameters using the Bayes approach aimed at maximizing the BLEU score. The Bayes approach uses the Gaussian process to model the process between parameters to optimize the probability of improvement [39]. Table 4 shows the number lists that were used in the hyper-parameter search.

| dim_model | tm_dim_ff | Batch_size |
| --- | --- | --- |
| 8, 16, 32, 64, 128, 256, 512, 1024, 2048 | 8, 16, 32, 64, 128, 256, 512, 1024, 2048 | 8, 16, 32, 64, 128 |

Table 4. Range of the hyper-parameter search for 'dim_model' (dimension model in Transformer), 'tm_dim_ff' (dimension in the feed forward layer), and the batch size

## 4. Results

### 4.1. BLEU Score

We used the BLEU score to evaluate the quality of training and to select the best models. The default hyperparameters used during the training of both models were (dim_model=512, tm_dim_ff=2048 and Batch size=64). The best parameter configurations by hyper-parameter search are (256, 2048, 16) and (512, 128, 32) for the En2Lu and Lu2En models, respectively, which achieve the best BLUE scores.

|  | En2Lu | Lu2En |
| --- | --- | --- |
| Valid (Before hyper-parameter search) | 16.13 | 20.59 |
| Valid (After hyper-parameter search) | 17.60 | 23.09 |
| Test | 17.47 | 21.28 |

Table 5. BLEU scores of translations (the test scores were measured from the model with the best hyper-parameter)

Table 5 shows that there is an improvement in the BLEU score of the validation set by +1.47 for En2Lu and +2.5 for Lu2En based on the hyper-parameter search. When we applied the model to the test dataset, the BLEU scores are 17.47 and 21.28 for En2Lu and Lu2En, respectively.

To understand the effect of hyper-parameters, we present the relevancy and correlation of the hyper-parameters to the prediction of the BLUE score in Figure 1. We used a sample configuration of 30 runs, each of which was a combination of 'dim_model', 'tm_dim_ff', and 'Batch_size'. For example, a combination (16, 32, 1024) makes a BLEU score of 13.96.

In Figure 1, we can see that there is a positive correlation between the BLEU score and the dim_model as well with tm_dim_ff, and a negative correlation with the batch size by the end of the hyper-parameter search. In addition, using Random Forest, the importance of the BLEU score was also calculated.

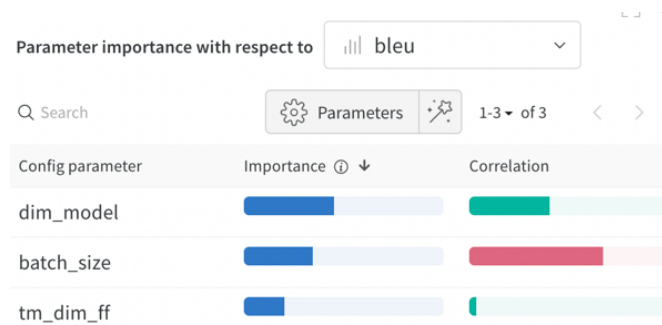

Figure 1. Correlation and importance of hyperparameters after training with wandb

The results indicate that for the En2Lu model, a moderately high dim_model value results in a high BLEU score, a low batch size results in a high BLEU score, and a slightly lower tm_dim_ff results in a slightly high BLEU score. Also, the dim_model is very important in determining the BLEU score, followed by the batch size and lastly tm_dim_ff.

## 4.2. Quality of Translation

After the selection of the best trained and validated model, it was applied to the test dataset to check the quality of translation. We selected some sample text from each of the languages. The first sentence as reflected in Tables 6 and 7 had a high translation quality meaning that it was not paraphrased by the model, whereas the second sentence had a low translation quality meaning that it was paraphrased by the model.

In both Tables 6 and 7, the same sentences are translated between both languages by the En2Lu and Lu2En models, respectively. The first sentence source (Src) in both tables is translated to the target language (Output) which is exactly the same as the target sentence (Trg). The translation of the second sentence is acceptable since the meaning of the output is quite similar to the meaning of the Trg sentence, though it is not exactly the same. The imperfect part in translation is underlined for each table. In Table 4, we have included the English translation of the Luganda output text in the parenthesis to show that the meaning is close to the Src sentence.

| Src (Eng) | Trg (Luganda) | Output (Luganda) |
|---|---|---|
| Mob Justice is highly Punished | *Okutwalira amateeka mu ngalo kibonerezebwa nnyo .* | Okutwalira amateeka mu ngalo kibonerezebwa nnyo . |
| People engaging in deforestation have been arrested | Abantu abeenyigira mu kusaanyawo ebibira bakwatiddwa | Abantu abeenyigira mu kutema emiti bakwatiddwa (***People engaging in cutting down trees have been arrested***) |

Table 6. Examples of translation from English to Luganda

| Src (Luganda) | Trg (Eng) | Output (Eng) |
|---|---|---|
| Okutwalira amateeka mu ngalo kibonerezebwa nnyo | Mob justice is punishable | Mob justice is punishable |
| Abantu abeenyigira mu kusaanyawo ebibira bakwatiddwa | People Engaging in deforestation have been arrested | People who engage in forestry activities have been arrested |

Table 7. Examples of translation from Luganda to English

## 5. Conclusion

Deep learning architectures require the use of very large datasets to train neural machine translation models. In this paper, we built a pairwise corpus of Luganda and English with a total of 41,070 sentences. We then trained NMT models by using the Transformer architecture with hyper-parameter search. The results indicate that the best trained model gives us a BLEU score of 17.47 for translation from English to Luganda and 21.28 from Luganda to English.

In the future, we intend to compare our translation with Google translate. By the time of submission of this paper, Google justified the need for the Luganda language and included it among the languages on the Google translate platform on May 11th, 2022.